\documentclass[11pt]{article}

\usepackage[preprint]{acl}

\usepackage{times}
\usepackage{latexsym}

\usepackage[T1]{fontenc}

\usepackage[utf8]{inputenc}

\usepackage{microtype}

\usepackage{inconsolata}
\usepackage{hyperref}
\usepackage{xurl}

\usepackage{graphicx}
\usepackage{algorithm}
\usepackage{algpseudocode}
\usepackage{amssymb}
\usepackage{amsmath}
\usepackage[table]{xcolor}
\usepackage{arydshln}
\usepackage{lipsum}
\usepackage{booktabs}
\usepackage{multirow}
\usepackage{listings}
\usepackage{xcolor}

\title{SigmaScale: LLM Compression with \\ SVD-based
Low-Rank Decomposition and Learned Scaling
Matrices}

\author{
 \textbf{Ernests Lavrinovics\textsuperscript{1}}\thanks{Worked performed as part of a research internship in CERN},
 \textbf{Marco Letizia\textsuperscript{2,3,4}},
 \textbf{Roy Janco\textsuperscript{5}},
 \textbf{Shai Segal}\thanks{Worked performed as part of Ceva Inc.},
\\
 \textbf{Johannes Bjerva\textsuperscript{1}},
 \textbf{Maurizio Pierini\textsuperscript{4}}, \\
 \textsuperscript{1}Department of Computer Science, Aalborg University Copenhagen, Denmark \\
 \textsuperscript{2}MaLGa-DIBRIS, University of Genoa, Genoa, Italy, \\
 \textsuperscript{3} INFN, Sezione di Genova, Genoa, Italy\\
 \textsuperscript{4}European Organization for Nuclear Research (CERN), Geneva, Switzerland\\
 \textsuperscript{5}Ceva, Inc.,
\\
 \small{
   \textbf{Correspondence:} \href{mailto:elav@cs.aau.dk}{elav@cs.aau.dk}
 }
}

\begin{document}
\maketitle

\begin{abstract}
We present SigmaScale, a method for learning auxiliary scaling matrices $S$ to aid truncated Singular Value Decomposition (SVD) based Large Language Model (LLM) compression. 
Instead of deriving scaling matrices analytically, SigmaScale optimizes two sets of vectors that define diagonal row and column scaling transformations under an activation-aware compression loss. 
We show that learned scaling lowers the effective intrinsic rank of weight matrices, as reflected by reductions in effective-rank entropy, and that this reduction is strongly correlated with compression loss.  
Experiments on Llama 3.1 8B Instruct and Qwen3-8B show that SigmaScale is competitive with closely related state-of-the-art SVD-based compression methods across perplexity and zero-shot benchmarks. 
By using learned activation-aware transformations, SigmaScale explores a more flexible route to low-rank LLM compression by adapting to the structure of individual model weights. The advantage observed in specific tasks makes our approach a valid option for applications requiring a reduced LLM-inference computing cost.
\end{abstract}

\section{Introduction and Background}

Large Language Models (LLMs) exhibit a remarkable performance and generalization across a variety of NLP tasks \cite{brown2020language} and it has been demonstrated that their performance scales with the increase of parameters \cite{kaplan2020scaling}, therefore leading to developments of very large language models in the tens and hundreds of billions of parameters \cite{grattafiori2024llama, deepseekai2026deepseekv4, yang2025qwen3}. The high parameter count impacts the technological accesibility and has significant environmental impacts due to the high power consumption of inference systems \cite{bommasani2021opportunities}. Therefore the AI research community has long explored methods of model compression \cite{zhu2024survey, liu2025survey} which span across quantization \cite{liu_spinquant_2025, ashkboos_quarot_2024, frantar_gptq_2023}, pruning  \cite{zhu_comprehensive_2025}, knowledge distillation (KD) \cite{yang2024llm, xin_quantization-aware_2026} and low-rank decomposition \cite{yuan2023asvd, wang2024svd, saha_compressing_2024}. Despite the success of these methods, practical deployment of quantization and pruning requires specialized hardware support which is a limitation contrary to low-rank decomposition and KD methods.

\begin{figure*}[!htb]
    \centering
    \includegraphics[width=1\linewidth]{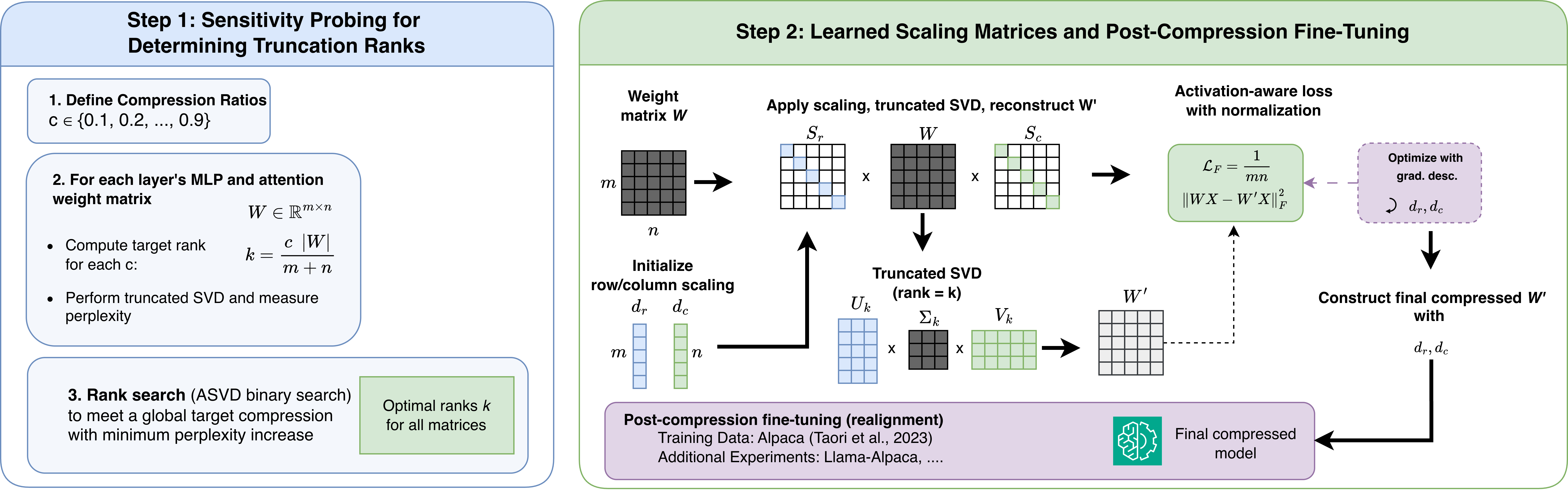}
    \caption{Visualization of the processing pipeline}
    \label{fig:pipeline}
\end{figure*}

Low-rank decomposition methods approximate a given matrix $W \in \mathbb{R}^{m \times n}$ as the product of two lower-rank matrices $L \in \mathbb{R}^{m \times k}$ and $R \in \mathbb{R}^{k \times n}$, where $k \ll \min(m, n)$. This means that low-rank decomposition typically does not require specialized hardware for supporting it, and it can be deployed alongside quantization and pruning \cite{yuan2023asvd, wang2024svd}. 

The Eckart–Young–Mirsky theorem \cite{eckart1936approximation, mirsky1960symmetric} states that, for minimizing the Frobenius norm $||W - W'||_F$, where $W$ is the original weight matrix and $W'$ is its low-rank approximation, the optimal analytical solution is given by the truncated singular value decomposition (SVD):
\begin{equation}
    f^{(k)}_{\rm svd}(W) = U_k \Sigma_k V_k^T = \sum_{i=1}^{k} u_i \sigma_i v_i^T.
\label{eq:truncated_svd}
\end{equation}
Here, $U_k \in \mathbb{R}^{m \times k}$ and $V_k \in \mathbb{R}^{n \times k}$ contain the top $k$ left and right singular vectors of $W$, respectively, while $\Sigma_k \in \mathbb{R}^{k \times k}$ is a diagonal matrix containing the corresponding $k$ largest singular values in descending order. Retaining only the top $k$ singular values and their corresponding singular vectors effectively discards components associated with lower-energy modes.
However, a drawback of SVD is its computational cost, $O(n^3)$ for square matrices \cite{shishkin2019fast, kishore2017literature}, and its unstable derivative, for which Taylor expansion-based approximations have been used to approximate its gradients \cite{wang_robust_2022, wang_dobi-svd_2025}. This means that performing SVD at each step of an optimization routine has its limitations, and it does not scale well as the matrix size increases.

Additionally naïve SVD decomposition on weight matrices $W$ minimizing the Frobenius norm $||W - W'||_F$ has been shown to perform poorly on neural network weight matrices \cite{Hsu2022LanguageMC, yuan2023asvd} partially due to the presence of outliers in the activations. Therefore, previous works \cite{pmlr-v119-nagel20a, wang2024svd, saha_compressing_2024} include the activations $x$ in the loss function $||Wx - W'x||_F$ to optimize over the functionality instead of the structure for a given weight matrix. Previous works further expand upon this idea by applying linear invertible scaling matrices $S$ to $W$ with the goal of: (1) absorbing outliers of the activations \cite{yuan2023asvd}, (2) aligning the singular values with the compression loss through Cholesky decomposition of the activation covariance matrix \cite{wang2024svd, li2026optimal}.

Since compression introduces a certain performance loss, compressed models are commonly fine-tuned to realign their weights. However, this is not straightforward for LLMs  primarily because these models undergo multi-step post-training.
Ideally, achieving a faithful distribution recovery after compression would require access to the same datasets used during the original post-training phases. In practice, this is often not achievable, as popular open-weight model technical reports \cite{grattafiori2024llama, yang2025qwen3} do not disclose the exact datasets employed during their post-training. To this end, KD \cite{hinton2015distilling} has been demonstrated to be useful for realigning the model to its original distribution \cite{xin_quantization-aware_2026}. Given that learning scaling matrices for improving the SVD performance is underexplored and previous methods \cite{yuan2023asvd, wang2024svd} rely on analytical means of deriving $S$, and given that KD is suggested to be beneficial for performance recovery over supervised fine-tuning, we cover the following contributions: (1) Empirical results on SVD compression performance when learning row- and column-wise scaling matrices. To the best of our knowledge, this is the first work to explore learning the parameters of scaling matrices $S$ for this purpose. (2) Comparisons between KD and supervised fine-tuning for performance recovery, with varied post-compression performance recovery datasets. (3) Custom variant of the Alpaca \cite{alpaca} dataset, based on Llama 3.1-8B Instruction output distribution. See Appendix \ref{sect:codebase_link} for the codebase link.

\section{Methodology}
The first step of our pipeline is sensitivity probing, which determines the compression levels for each given layer and module of the model, described in Section \ref{sect:sens_probing}. The second step is to learn scaling matrices that apply a linear transformation to the weight matrix $W$ before performing truncated SVD. After the optimal scaling matrix has been learned, we perform the final compression on the model and do post-compression fine-tuning for realignment of weights. We base our experiments on Llama 3.1 8B-Instruct \cite{grattafiori2024llama} and Qwen3-8B models \cite{yang2025qwen3}. See Figure \ref{fig:pipeline} for our pipeline visualization.

\subsection{Sensitivity Probing for Determining Truncation Ranks}
\label{sect:sens_probing}
Sensitivity probing is done by defining a set of compression ratios $c \in \{0.1, 0.2, \dots, 0.9\}$ which are used to calculate the truncated SVD target rank $k$ using Eq \ref{eq:compression_ratios_svd}. Intuitively the compression ratios describe the percentage of the parameter count that will be retained after the decomposition. 
\begin{equation}
k =
c\,|\mathbf{W}| \left(m + n\right)^{-1}.
\label{eq:compression_ratios_svd}
\end{equation}
Where $c$ denotes the compression ratio, $|\mathbf{W}|$ the number of parameters in the weight matrix, and $m, n$ are the dimensions of $W$ rows and columns.

We probe for perplexity metric in our condition models by performing truncated SVD compression at rank $k$ for each isolated MLP and attention weight matrix at each layer. This information is used to find most optimal set of compression ranks $k$ across the whole model that achieve the global target compression ratio, while minimizing the increase in perplexity. This rank search is done with the binary search algorithm introduced in ASVD \cite{yuan2023asvd}. 
Truncation is performed by retaining the first $k$ singular values and discarding the tail-end of the distribution.

\subsection{Learned Scaling Matrices and Post Compression Fine-Tuning}
For a given weight matrix $W \in \mathbb{R}^{m \times n}$, we initialize two vectors $d_r \in \mathbb{R}^{m}, d_c \in \mathbb{R}^{n}$ with a scaled Gaussian distribution:
$ d_{r,c}=(0.1)\,\sigma_W\,\epsilon_{r,c}$ with $\epsilon_r\sim\mathcal{N}(0,I_m)$ and $\epsilon_c\sim\mathcal{N}(0,I_n)$. We use the standard deviation $\sigma_W$ of the weight matrix to scale the initialization of $d_r$ and $d_c$, to match the $d_r$ and $d_c$ with the scaled magnitude of the corresponding weight matrix.

From the vectors $d_r$ and $d_c$, we construct positive diagonal scaling via exponentiation, defined as $S_r = \rm{diag}(\exp(d_r))$ and $S_c = \rm{diag}(\exp(d_c)$. These are used to apply row and column scaling to model weights $W$. We then perform truncated SVD (Eq \ref{eq:truncated_svd}), and apply the inverse scaling (Eq. \ref{eq:compression_eq}) before computing an activation aware loss with a normalization term (Eq. \ref{eq:act_aware_loss}).
\begin{equation}
W' =
S_r^{-1}
f^{(k)}_{\mathrm{svd}}(S_r W S_c)
S_c^{-1}
\label{eq:compression_eq}
\end{equation}
\begin{equation}
\mathcal{L}_{\mathrm{F}} =
\frac{1}{mn}
\left\|
W X - W' X
\right\|_F^2.
\label{eq:act_aware_loss}
\end{equation}
Here, $W$ is the original weight matrix, $X$ are activations from a calibration set, $W'$ is the compressed weight matrix.

After learning $d_r$ and $d_c$, we construct the final compressed weight matrix $W'$ and replace the original matrix in the model. We first apply truncated SVD to the scaled weight matrix: 
$f^{(k)}_{\mathrm{svd}}(S_r W S_c)$
The final low-rank factors are then obtained by absorbing the singular values and applying the inverse scaling transformations:
\begin{equation}
L =
S_r^{-1} U_k \sqrt{\Sigma_k},
\qquad
R =
\sqrt{\Sigma_k} V_k^T S_c^{-1},
\label{eq:reconstruction}
\end{equation}
such that the compressed matrix satisfies $W' = LR$. Finally, post-compression fine-tuning is performed to realign the impaired weight matrices. See Appendix \ref{appx:impl_details} for further details including pseudo-code.

\section{Experimental setup}
As part of the experiments, we use Qwen3-8B and Llama 3.1-8B-Instruction models with a focus on English language. We use a Wikitext2-raw-v1 \cite{merity2016pointer} test split with n=141 samples and 2048 sequence length for all perplexity measurements. As our calibration data, we use a set of n=32 samples of 2048 sequence length from Wikitext training split. Alpaca \cite{alpaca} is used for post-compression fine-tuning. See Appendix \ref{appx:impl_details} for a full set of implementation details. Evaluation is done on five downstream task benchmarks with licensing terms summarized in Appendix \ref{appx:licensing}. Our compute budget is described in Appendix \ref{appx:compute_budget}.

During post-compression fine-tuning we freeze all weight matrices that have not been modified by the low-rank decomposition and perform comparisons with supervised fine-tuning versus knowledge distillation (KD) using an uncompressed teacher model. Our experimental setup does not perform compression on token embeddings, layer normalizations or language modeling head. We run comparisons with SVD-LLM \cite{wang2024svd} and ASVD+ \cite{yuan2023asvd} for which we unify the hyperparameter sets for direct comparisons and perform supervised-fine-tuning for performance recovery with frozen, non-compressed elements of the model. We use LM-Evaluation-Harness framework for running evaluations \cite{eval-harness} on full downstream task benchmarks.

\begin{table*}[!htb]
\scriptsize
\centering
\begin{tabular}{lllcp{1,75cm}p{1,75cm}p{1,75cm}p{1,75cm}p{1,75cm}p{1,75cm}}
\toprule
& Compression & Method & PPL$\downarrow$
& OpenBookQA \cite{mihaylov2018can}
& ARC-Easy \cite{allenai_arc}
& WinoGrande \cite{sakaguchi2021winogrande}
& PIQA \cite{Bisk2020}
& HellaSwag \cite{zellers2019hellaswag} \\
&&&
& Acc$_{\text{norm}}$
& Acc$_{\text{norm}}$
& Acc$_{\text{norm}}$
& Acc$_{\text{norm}}$
& Acc$_{\text{norm}}$ \\
\midrule

\multirow{13}{*}{\rotatebox{90}{\textbf{Llama 3.1 8B Instruct}}} 
& 1.00x & Baseline & 7.21
& 43.00 $\pm$ 2.22
& 79.63 $\pm$ 0.83
& 78.06 $\pm$ 1.16
& 80.96 $\pm$ 0.92
& 80.12 $\pm$ 0.40 \\

\cmidrule{2-9} 
& 0.90x & SVD-LLM & 13.31
& 37.80 $\pm$ 2.17
& 68.10 $\pm$ 0.96
& 65.98 $\pm$ 1.33
& 75.24 $\pm$ 1.01
& 64.71 $\pm$ 0.48 \\

& 0.90x & ASVD+ & \textbf{8.26}
& 42.60 $\pm$ 2.21
& 76.89 $\pm$ 0.86
& 70.64 $\pm$ 1.28
& 79.49 $\pm$ 0.94
& 74.35 $\pm$ 0.44 \\

& 0.90x & SigmaScale & 8.95
& \textbf{43.00 $\pm$ 2.22}
& \textbf{78.62 $\pm$ 0.84}
& 73.32 $\pm$ 1.24
& \textbf{79.54 $\pm$ 0.94}
& 75.98 $\pm$ 0.43 \\

& 0.90x & SigmaScale (KD) & 8.70
& 42.80 $\pm$ 2.21
& 77.86 $\pm$ 0.85
& \textbf{73.72 $\pm$ 1.24}
& 79.38 $\pm$ 0.94
& \textbf{76.14 $\pm$ 0.43} \\

\cmidrule{2-9}
& 0.75x & SVD-LLM & 18.15
& 32.60 $\pm$ 2.10
& 60.65 $\pm$ 1.00
& 60.06 $\pm$ 1.38
& 70.67 $\pm$ 1.06
& 53.84 $\pm$ 0.50 \\

& 0.75x & ASVD+ & \textbf{13.67}
& 33.40 $\pm$ 2.11
& 55.93 $\pm$ 1.02
& 59.27 $\pm$ 1.38
& 69.53 $\pm$ 1.07
& 56.14 $\pm$ 0.50 \\

& 0.75x & SigmaScale & 18.48
& 36.80 $\pm$ 2.16
& 63.89 $\pm$ 0.99
& 62.43 $\pm$ 1.36
& \textbf{73.78 $\pm$ 1.03}
& 61.41 $\pm$ 0.49 \\

& 0.75x & SigmaScale (KD) & 17.90
& \textbf{37.00 $\pm$ 2.16}
& \textbf{64.52 $\pm$ 0.98}
& \textbf{64.09 $\pm$ 1.35}
& 73.72 $\pm$ 1.03
& \textbf{61.50 $\pm$ 0.49} \\

\cmidrule{2-9}
& 0.50x & SVD-LLM & \textbf{39.83}
& 27.20 $\pm$ 1.99
& \textbf{45.29 $\pm$ 1.02}
& 51.14 $\pm$ 1.40
& \textbf{62.13 $\pm$ 1.13}
& \textbf{36.30 $\pm$ 0.48} \\

& 0.50x & ASVD+ & 48.39
& \textbf{30.00 $\pm$ 2.05}
& 34.39 $\pm$ 0.97
& \textbf{51.85 $\pm$ 1.40}
& 57.89 $\pm$ 1.15
& 33.26 $\pm$ 0.47 \\

& 0.50x & SigmaScale & 138.63
& 26.80 $\pm$ 1.98
& 41.08 $\pm$ 1.01
& 50.91 $\pm$ 1.41
& 61.59 $\pm$ 1.13
& 33.62 $\pm$ 0.47 \\

& 0.50x & SigmaScale (KD) & 121.85
& 28.00 $\pm$ 2.01
& 41.46 $\pm$ 1.01
& 51.78 $\pm$ 1.40
& 61.81 $\pm$ 1.13
& 33.98 $\pm$ 0.47 \\

\midrule

\multirow{13}{*}{\rotatebox{90}{\textbf{Qwen3-8B}}}
& 1.00x & Baseline & 9.72
& 41.40 $\pm$ 2.20
& 80.93 $\pm$ 0.81
& 70.80 $\pm$ 1.28
& 77.69 $\pm$ 0.97
& 76.49 $\pm$ 0.42 \\

\cmidrule{2-9}
& 0.90x & SVD-LLM & 11.51
& \textbf{43.00 $\pm$ 2.22}
& 74.66 $\pm$ 0.89
& \textbf{69.85 $\pm$ 1.29}
& 75.57 $\pm$ 1.00
& 68.33 $\pm$ 0.46 \\

& 0.90x & ASVD+ & \textbf{10.11}
& 36.00 $\pm$ 2.15
& 74.62 $\pm$ 0.89
& 63.30 $\pm$ 1.35
& 76.01 $\pm$ 1.00
& 65.41 $\pm$ 0.47 \\

& 0.90x & SigmaScale & 10.89
& 40.80 $\pm$ 2.20
& \textbf{80.18} $\pm$ 0.82
& 65.90 $\pm$ 1.33
& \textbf{77.75} $\pm$ 0.97
& 68.09 $\pm$ 0.47 \\

& 0.90x & SigmaScale (KD) & 10.84
& 40.40 $\pm$ 2.20
& 79.63 $\pm$ 0.83
& 65.82 $\pm$ 1.33
& 77.64 $\pm$ 0.97
& \textbf{68.49} $\pm$ 0.46 \\



\cmidrule{2-9}
& 0.75x & SVD-LLM & 13.64
& \textbf{40.80 $\pm$ 2.20}
& 71.42 $\pm$ 0.93
& \textbf{67.17 $\pm$ 1.32}
& 72.69 $\pm$ 1.04
& \textbf{62.48 $\pm$ 0.48} \\

& 0.75x & ASVD+ & \textbf{12.34}
& 36.20 $\pm$ 2.15
& 64.77 $\pm$ 0.98
& 59.19 $\pm$ 1.38
& 71.98 $\pm$ 1.05
& 58.71 $\pm$ 0.49 \\

& 0.75x & SigmaScale & 14.68
& 40.40 $\pm$ 2.20
& 74.28 $\pm$ 0.90
& 59.98 $\pm$ 1.38
& \textbf{73.94} $\pm$ 1.02
& 58.81 $\pm$ 0.49 \\

& 0.75x & SigmaScale (KD) & 14.43
& 40.00 $\pm$ 2.19
& \textbf{75.46} $\pm$ 0.88
& 60.85 $\pm$ 1.37
& 73.88 $\pm$ 1.02
& 59.16 $\pm$ 0.49 \\



\cmidrule{2-9}
& 0.50x & SVD-LLM & \textbf{21.84}
& \textbf{36.80 $\pm$ 2.16}
& 55.51 $\pm$ 1.02
& \textbf{62.12 $\pm$ 1.36}
& \textbf{66.76 $\pm$ 1.10}
& \textbf{47.66 $\pm$ 0.50} \\

& 0.50x & ASVD+ & 24.30
& 29.80 $\pm$ 2.05
& 39.27 $\pm$ 1.00
& 55.17 $\pm$ 1.40
& 59.30 $\pm$ 1.15
& 39.49 $\pm$ 0.49 \\

& 0.50x & SigmaScale & 31.92
& 32.00 $\pm$ 2.03
& 57.00 $\pm$ 1.02
& 54.85 $\pm$ 1.40
& 65.07 $\pm$ 1.11
& 39.40 $\pm$ 0.49 \\

& 0.50x & SigmaScale (KD) & 31.29
& 32.60 $\pm$ 2.10
& \textbf{57.62} $\pm$ 1.01
& 54.38 $\pm$ 1.40
& 64.69 $\pm$ 1.12
& 39.48 $\pm$ 0.49 \\


\bottomrule
\end{tabular}
\caption{Post-compression fine tuning results for Llama 3.1 8B Instruct and Qwen3-8B. Zero-shot benchmarks report length-normalized accuracy with standard error, \textit{ppl} reports mean perplexity over Wikitext-Test split.}
\label{tab:llama_qwen3_results}
\end{table*}

\section{Results and Analysis}
\label{sect:results}
Table \ref{tab:llama_qwen3_results} show results for Llama 3.1-8B-Instruction and Qwen3-8B models with SigmaScale comparisons in KD and supervised fine-tuning paradigms. 
SigmaScale is most competitive in the mild-to-moderate compression regime. At 0.90x retention, it substantially improves perplexity over SVD-LLM for both models, while also recovering much of the zero-shot performance. 
At 0.75x retention SigmaScale generally improves several zero-shot benchmarks, but perplexity gains are marginal.

At 0.50x retention, SigmaScale degrades more sharply, particularly for Llama 3.1-8B-Instruction. 
This suggests that the method is most effective when reshaping the singular-value spectrum can preserve the dominant components of the weight matrix. Under aggressive compression, the retained subspace may become too small for learned scaling alone to compensate for the discarded singular directions. SigmaScale should therefore be understood as a mechanism for improving truncation quality in the retained-rank regime, rather than as a complete solution for extreme low-rank compression. Contrary to \cite{xin_quantization-aware_2026}, our results do not show major improvements of KD over supervised fine-tuning conditions for SigmaScale.

Given that singular values are indicative of the intrinsic rank \cite{konstantinides2002statistical}, we perform an analysis of the given compressed weight matrices during the optimization of scaling vectors $d_{r,c}$. In Table \ref{tab:module_loss_entr_corr} we aggregate the mean drop in compression loss as per Eq. \ref{eq:act_aware_loss} and also measure the average drop in the effective rank entropy \cite{roy2007effective} of the $\Sigma$ component. We see that there is a strong correlation between the compression loss and the effective rank entropy of the compressed weight matrices' $\Sigma$ components. See Appendix \ref{appx:sigma_plots} for further visualizations and corresponding results for Qwen3 model, for which we observe similar patterns.

\begin{table}[!htb]
    \small
    \centering
    \begin{tabular}{llll}
    \toprule
        Module & Loss $\Delta$ \% & Entropy $\Delta$ \% & Corr \\
        \midrule
        mlp\_down\_proj & -22.418 & -0.302 & 0.923 \\ 
        mlp\_gate\_proj & -31.145 & -2.225 & 0.916 \\
        mlp\_up\_proj & -33.73 & -1.796 & 0.919 \\ 
        self\_attn\_k\_proj & -44.278 & -6.272 & 0.814 \\
        self\_attn\_o\_proj & -20.786 & -4.895 & 0.917 \\
        self\_attn\_q\_proj & -32.366 & -8.542 & 0.857 \\
        self\_attn\_v\_proj & -33.68 & -2.492 & 0.886 \\
        \bottomrule
    \end{tabular}
    \caption{Llama 3.1 average percentage of loss and effective rank entropy decrease during scaling matrix training}
    \label{tab:module_loss_entr_corr}
\end{table}

\section{Conclusions}
Our work demonstrates the effectiveness of learning scaling matrices $S$ for SVD-based LLM compression. Our results show that SigmaScale performs on par with the most similar state-of-the-art methods, while taking a fundamentally different approach: learning $S$ rather than deriving it analytically, as in SVD-LLM or ASVD. We show that the learned scaling matrices manipulate the intrinsic rank of a given weight matrix, as reflected by changes in the effective-rank entropy of the singular values and its correlation with compression loss. Future work should further investigate the impact of calibration data used to learn $S$, explore different initialization strategies for $S$, and examine how complementary current state-of-the-art methods are to one another.

\section*{Limitations}
Our method relies on computing SVD at every update step while learning the scaling matrix $S$ which has $O(n^3)$ computational expense, we do not explore faster alternative SVD methods that would use approximations.

Our method, as shown in Section \ref{sect:results} degrades sharply (especially for Llama 3.1) and should not be viewed as a complete solution for extreme low-rank compression. Our evaluation is based on perplexity and a specific set of zero-shot benchmarks. We do not explore effects on longer-form generation, or coding tasks.  

Current method's robustness to different calibration distributions has not been formally verified, yet we anticipate that at its core, Wikitext is a subpar choice which was used mainly to stay consistent for comparisons with SVD-LLM and ASVD.

\section*{Ethical Considerations}
To the best of our knowledge, our work does not require an additional ethics review.  We do not conduct tests on humans nor use any sensitive data. We summarize used asset licenses in Appendix \ref{appx:licensing} for which our work does not violate any of the licensing terms. We do not foresee additional significant ethical, societal, or environmental risks arising directly from this work. As common in the field, we urge anyone who uses our work for downstream applications to cross-check and verify their model integrity before production deployments.


\bibliography{custom}

\appendix

\section{Core Experiment Variations}
\label{sect:appx_experiment_variations}

\begin{table*}[!tb]
\scriptsize
\begin{tabular}{cllcccccccccc}
\toprule
& & & & \multicolumn{1}{c}{OpenBookQA} & \multicolumn{1}{c}{ARC-Easy} & \multicolumn{1}{c}{WinoGrande} & \multicolumn{1}{c}{PIQA} & \multicolumn{1}{c}{HellaSwag} \\

\cmidrule(lr){5-5} \cmidrule(lr){6-6} \cmidrule(lr){7-7} \cmidrule(lr){8-8} \cmidrule(lr){9-9}
& Compression & Method & PPL$\downarrow$
& Acc$_{\text{norm}}$
& Acc$_{\text{norm}}$
& Acc
& Acc$_{\text{norm}}$
& Acc$_{\text{norm}}$ \\
\midrule

\multirow{15}{*}{\rotatebox[origin=c]{90}{\textbf{Llama 3.1 8B Instruct}}} 
& 1.00x & Baseline & 7.21
& 43.00 $\pm$ 2.22
& 79.63 $\pm$ 0.83
& 78.06 $\pm$ 1.16
& 80.96 $\pm$ 0.92
& 80.12 $\pm$ 0.40 \\
\cmidrule{2-9} 

& 0.90x & SigmaScale ($\heartsuit$) & 9.29
& \textbf{44.00} $\pm$ 2.22
& 77.40 $\pm$ 0.86
& 72.69 $\pm$ 1.25
& 79.00 $\pm$ 0.95
& \textbf{78.32} $\pm$ 0.41 \\

& 0.90x & SigmaScale ($\heartsuit$, KD=0.7) & 8.55
& 43.80 $\pm$ 2.22
& \textbf{78.20} $\pm$ 0.85
& 74.82 $\pm$ 1.22
& 79.05 $\pm$ 0.95
& 76.76 $\pm$ 0.42 \\

& 0.90x & SigmaScale ($\clubsuit$) & 8.81
& 42.40 $\pm$ 2.21
& 76.98 $\pm$ 0.86
& 73.40 $\pm$ 1.24
& 78.89 $\pm$ 0.95
& 76.68 $\pm$ 0.42 \\

& 0.90x & SigmaScale ($\clubsuit$ KD=0.7) & 8.51
& 43.00 $\pm$ 2.22
& 76.73 $\pm$ 0.87
& 74.11 $\pm$ 1.23
& 78.94 $\pm$ 0.95
& 76.51 $\pm$ 0.42 \\

& 0.90x & SigmaScale ($\clubsuit$ KD=1) & \textbf{8.44}
& 42.80 $\pm$ 2.21
& 76.60 $\pm$ 0.87
& \textbf{74.98} $\pm$ 1.22
& \textbf{79.16} $\pm$ 0.95
& 76.58 $\pm$ 0.42 \\

& 0.90x & SigmaScale ($\spadesuit$) & 9.03
& 42.00 $\pm$ 2.21
& 76.14 $\pm$ 0.87
& 72.69 $\pm$ 1.25
& 79.00 $\pm$ 0.95
& 77.45 $\pm$ 0.42 \\

& 0.90x & SigmaScale ($\spadesuit$ KD) & 8.55
& 42.20 $\pm$ 2.21
& 77.19 $\pm$ 0.86
& 74.90 $\pm$ 1.22
& \textbf{79.16} $\pm$ 0.95
& 77.08 $\pm$ 0.42 \\

\cmidrule{2-9} 

& 0.75x & SigmaScale ($\heartsuit$) & 22.42
& 37.60 $\pm$ 2.17
& 64.98 $\pm$ 0.98
& 62.43 $\pm$ 1.36
& 74.10 $\pm$ 1.02
& 64.72 $\pm$ 0.48 \\

& 0.75x & SigmaScale ($\heartsuit$, KD=0.7) & 17.10
& 39.00 $\pm$ 2.18
& 65.87 $\pm$ 0.97
& \textbf{65.75} $\pm$ 1.33
& \textbf{74.54} $\pm$ 1.02
& 64.04 $\pm$ 0.48 \\

& 0.75x & SigmaScale ($\clubsuit$) & 17.03
& 38.60 $\pm$ 2.18
& 65.61 $\pm$ 0.97
& 65.04 $\pm$ 1.34
& 74.05 $\pm$ 1.02
& 64.91 $\pm$ 0.48 \\

& 0.75x & SigmaScale ($\clubsuit$, KD 0.7) & 15.97
& 38.40 $\pm$ 2.18
& \textbf{66.08} $\pm$ 0.97
& 65.19 $\pm$ 1.34
& 74.21 $\pm$ 1.02
& 64.65 $\pm$ 0.48 \\

& 0.75x & SigmaScale ($\clubsuit$ KD=1) & \textbf{15.71}
& 38.20 $\pm$ 2.18
& 65.40 $\pm$ 0.98
& 65.67 $\pm$ 1.33
& 74.37 $\pm$ 1.02
& 64.28 $\pm$ 0.48 \\

& 0.75x & SigmaScale ($\spadesuit$ 1para, 3epoch) & 19.89
& \textbf{39.40} $\pm$ 2.19
& 63.64 $\pm$ 0.99
& 64.09 $\pm$ 1.35
& 73.34 $\pm$ 1.03
& \textbf{66.02} $\pm$ 0.47 \\

& 0.75x & SigmaScale ($\spadesuit$ KD=0.7) & 16.74
& 38.60 $\pm$ 2.18
& 65.95 $\pm$ 0.97
& 64.96 $\pm$ 1.34
& 74.21 $\pm$ 1.02
& 65.76 $\pm$ 0.47 \\

\bottomrule
\end{tabular}
\caption{Llama-Alpaca variations: $\clubsuit$ 3 para, 1 epoch; $\spadesuit$ 1para 3 epoch. $\heartsuit$ Vanilla Alpaca with 3 training epochs}
\label{tab:llama_alpaca_results}
\end{table*}

 \begin{table*}[t]
\centering
\setlength{\tabcolsep}{4pt}
\scriptsize
\begin{tabular}{llcccccc}
\toprule
& & & OpenBookQA & ARC-Easy & WinoGrande & PIQA & HellaSwag \\
\cmidrule(lr){4-4} \cmidrule(lr){5-5} \cmidrule(lr){6-6} \cmidrule(lr){7-7} \cmidrule(lr){8-8}
Compression & Method & PPL$\downarrow$
& Acc$_{\text{norm}}$
& Acc$_{\text{norm}}$
& Acc$_{\text{norm}}$
& Acc$_{\text{norm}}$
& Acc$_{\text{norm}}$ \\
\midrule

1.00x & Baseline & 7.21
& 43.00 $\pm$ 2.22
& 79.63 $\pm$ 0.83
& 78.06 $\pm$ 1.16
& 80.96 $\pm$ 0.92
& 80.12 $\pm$ 0.40 \\
\midrule

0.90x & SigmaScale & 7.89
& 42.20 $\pm$ 2.21
& 77.57 $\pm$ 0.86
& 71.90 $\pm$ 1.26
& 78.73 $\pm$ 0.95
& 75.83 $\pm$ 0.43 \\

0.90x & SigmaScale (KD) & 8.08
& 42.20 $\pm$ 2.21
& 78.07 $\pm$ 0.85
& 72.93 $\pm$ 1.25
& 78.62 $\pm$ 0.96
& 75.86 $\pm$ 0.43 \\

\midrule

0.75x & SigmaScale & 14.18
& 32.60 $\pm$ 2.10
& 54.76 $\pm$ 1.02
& 60.46 $\pm$ 1.37
& 70.89 $\pm$ 1.06
& 54.68 $\pm$ 0.50 \\

0.75x & SigmaScale (KD) & 14.07
& 33.80 $\pm$ 2.12
& 56.82 $\pm$ 1.02
& 61.33 $\pm$ 1.37
& 70.84 $\pm$ 1.06
& 55.82 $\pm$ 0.50 \\

\bottomrule
\end{tabular}

\caption{Benchmark results for Wikitext-Train as post-compression training data for Llama 3.1 8B Instruct model.}
\label{tab:sigma_scale_wikitext}
\end{table*} 

We perform additional experiments with a custom Alpaca dataset for which its outputs are generated from Llama 3.1 8B Instruction model. The dataset contains three output generations per single instruction, with a goal to introduce variance. We perform post-compression fine-tuning by training on 1 answer per instruction over 3 epochs against 3 answers per instruction over 1 epoch. The results are depicted in Table \ref{tab:llama_alpaca_results} and our tests show minor improvements with Llama-Alpaca dataset. Specifically for 25\% compression, there is 1 point perplexity improvement for between $\clubsuit$ and $\heartsuit$ experiment variations but marginal changes across zero-shot benchmarks. Further details of the custom Alpaca dataset are described in Section \ref{appx:custom_alpaca_dataset_card}.

We also run experiments for using Wikitext2 as post-compression fine-tuning dataset by training on the continued pretraining task. Results for this are depicted in Table \ref{tab:sigma_scale_wikitext} which showcases improvements in perplexity although decrease in overall zero-shot benchmark performance across the board.

\section{Implementation Details}
\label{appx:impl_details}

\textbf{Data:}
For Wikitext subsplits we use Wikitext2-raw-v1 \footnote{\url{huggingface.co/datasets/Salesforce/wikitext/viewer/wikitext-2-raw-v1}}.
Calibration data: Wikitext2-raw-v1-Train for Llama 3.1 and Qwen3. Original Alpaca \footnote{\url{huggingface.co/datasets/tatsu-lab/alpaca}} \cite{alpaca} for post-compression fine-tuning. Wikitext-Test for all perplexity evaluations.

\textbf{Scaling Matrix Learning:} When learning the row and column scaling matrices, we perform hyperparameter optimization via grid search. Optimal found configuration is described in Table \ref{tab:hparam_sweep}. See Algorithm \ref{alg:svd_algo} for pseudo-code of the training loop and Algorithm \ref{alg:reconstruction_algo} for pseudo-code of constructing final compressed $W'$ .

\begin{table}[H]
\small
\centering
\begin{tabular}{ll}
\hline
\textbf{Hyperparameter} & \textbf{Values} \\
\hline
\texttt{Gaussian scale} & $\{10^{-5}, \mathbf{10^{-6}}, 10^{-7}, \texttt{None}\}$ \\
\texttt{Weight decay} & $\{10^{-3}, 10^{-4}, 10^{-5}, \mathbf{10^{-6}}\}$ \\
\texttt{learning rate(LR)} & $\{0.001, \textbf{0.005}, 0.0005, 10^{-5}\}$ \\
\texttt{LR scheduler} & $\{\texttt{cosine}, \texttt{plateau}, \texttt{\textbf{none}}\}$ \\
\hline
\end{tabular}
\caption{Hyperparameter sweep configuration}
\label{tab:hparam_sweep}
\end{table}

\textbf{Post-compression fine-tuning:} We perform post-compression fine-tuning with full Alpaca dataset training split over \textbf{1 epoch} and computing the loss only over the response span. Wikitext-Test for all perplexity evaluations is first tokenized then split into 2048 sequences. SVD-LLM and ASVD has a discrepancy where first their implementations split text chunks into $sequence\_len \times 10$ character lengths and afterwards perform tokenization. We use the same learning rate and epoch count for SigmaScale, ASVD and SVD-LLM.

Our post-compression fine-tuning uses Alpaca dataset \cite{alpaca} for which we fine-tune over 1 epoch computing cross-entropy loss over the response span. We use learning rate $10^{-6}$ with a cosine LR scheduler with 0.1 warmup step ratio, this configuration is used for both SVD-LLM and SigmaScale results.

\textbf{Knowledge distillation (KD):} We use the following loss function (Eq. \ref{eq:kd}) for performing KD

\begin{equation}
\mathcal{L}_{\text{total}} = \alpha \mathcal{L}_{\text{KD}} + (1 - \alpha) \mathcal{L}_{\text{task}}
\label{eq:kd}
\end{equation}
where $\mathcal{L}_{\text{KD}}$ is the KL divergence between student and teacher logits and $\mathcal{L}_{\text{task}}$ is cross-entropy of student predictions over ground truth labels. By default we always use $\alpha=0.7$ unless explicitly specified otherwise.

\begin{algorithm*}[!htb]
\begin{algorithmic}[1]
\State \textbf{Input:} Matrix $W^{m \times n}$, rank $k$, number of epochs $T$, activations $X$ from calibration data
\State Initialize $d_c \sim \mathcal{N}(0,I_n) \cdot \sigma_w \cdot 0.1$; $d_r \sim \mathcal{N}(0,I_m) \cdot \sigma_w \cdot 0.1$

\For{$t = 0$ to $T-1$}

    \Comment{\textit{Construct the scaling matrices and their inversions}}
    \State $S_c \gets \mathrm{diag}(exp(d_c))$
    \State $S_c^{-1} \gets \mathrm{diag}(exp(-d_c))$

    \State $S_r \gets \mathrm{diag}(exp(d_r))$
    \State $S_r^{-1} \gets \mathrm{diag}(exp(-d_r))$

    \Comment{\textit{Apply column and row scaling}}
    \State $W_{\text{scaled}} \gets S_r W S_c$

    \Comment{\textit{Compute truncated SVD}}
    \State $(U_k, S_k, V_k) \gets \mathrm{SVD}(W_{\text{scaled}}, k)$

    \Comment{\textit{Reconstruct W' truncated SVD}}
    \State $W_{\text{scaled}}^{(k)} \gets U_k \mathrm{diag}(S_k) V_k$

    \Comment{\textit{Invert scaling}}
    \State $W^{(r)} \gets S_r^{-1} W_{\text{scaled}}^{(r)} S_c^{-1}$

    \Comment{ ... Compute loss, update $d_c$ $d_r$}

\EndFor
\State \textbf{Output:} $d_c, d_r$

\end{algorithmic}
\caption{Pseudo-code of training the scaling matrix $S$}
\label{alg:svd_algo}
\end{algorithm*}

\textbf{SVD derivative during training:} As outlined in contribution \cite{wang_robust_2022}, the SVD algorithm has an unstable derivative, we bypass this by skipping update steps in which the $\sigma_i - \sigma_j$ denominator reaches close to 0 causing NaN values. We find that even with the skipped updates, our loss still converges often triggering early stop, therefore while this is not necessarily a robust solution, we do not experience this as a bottleneck for our usecase.

\section{Compute Budget}
\label{appx:compute_budget}
For running our computation we use Nvidia and AMD GPUs summarized in Table \ref{tab:compute_times} with approximate compute times and GPU count used for a given processing stage. Numbers are reported per experimental condition (e.g. model and corresponding compression ratio).

\begin{table}[!htb]
\small
\begin{tabular}{p{2,5cm}p{2cm}p{2cm}}
\toprule
Stage                        & Average mean time & GPUs           \\
\midrule
Scaling Matrix Training      & 45,5h                 & Nvidia H100 x2 \\
Post-compression fine-tuning & 2h                    & AMD MI300X x2 \\             
\bottomrule
\end{tabular}
\caption{Compute budget overview}
\label{tab:compute_times}
\end{table}

\section{Custom Alpaca Dataset}
\label{appx:custom_alpaca_dataset_card}
Created with Llama 3.1 8B Instruct by generating 3 output versions per single datapoint row. The idea is to introduce data variance for weight realignment. For creating the dataset, we re-ran the inference 3 times with the following generation settings Table \ref{tab:alpaca_ds_gen_parmas}.

\begin{table}[!htb]
\centering
\begin{tabular}{ll}
\hline
\textbf{Parameter} & \textbf{Value} \\
\hline
\texttt{max\_new\_tokens} & 1024 \\
\texttt{temperature} & 0.7 \\
\texttt{top\_p} & 0.9 \\
\texttt{do\_sample} & True \\
\texttt{repetition\_penalty} & 1.1 \\
\texttt{eos\_token\_id} & terminators \\
\texttt{pad\_token\_id} & tokenizer.pad\_token\_id \\
\texttt{use\_cache} & True \\
\hline
\end{tabular}
\caption{Hyperparameters used for generating Llama 3.1 8B answers of the Alpaca inputs for a custom dataset used in experiments described in Table \ref{tab:llama_alpaca_results}.}
\label{tab:alpaca_ds_gen_parmas}
\end{table}

\begin{algorithm*}[!htb]
\begin{algorithmic}[1]

\State \textbf{Input:} Weight matrix $W$, rank $r$, scaling vectors $d_r, d_c$

\Comment{\textit{Construct scaling matrices and their inverses}}
\State $S_c \gets \mathrm{diag}(\exp(d_c))$
\State $S_c^{-1} \gets \mathrm{diag}(\exp(-d_c))$
\State $S_r \gets \mathrm{diag}(\exp(d_r))$
\State $S_r^{-1} \gets \mathrm{diag}(\exp(-d_r))$

\Comment{\textit{Apply symmetric row/column scaling}}
\State $W_{\mathrm{scaled}} \gets S_r W S_c$

\Comment{\textit{Compute truncated SVD}}
\State $(U_k, \Sigma_k, V_k) \gets \mathrm{TruncateSVD}(W_{\mathrm{scaled}}, k)$

\Comment{\textit{Construct square-root singular value factors}}
\State $L_{\text{scaled}} \gets U_k \sqrt{\Sigma_k}$
\State $R_{\text{scaled}} \gets \sqrt{\Sigma_k} V_k^\top$

\Comment{\textit{Map factors back to original parameter space}}
\State $L \gets S_r^{-1} L_{\text{scaled}}$
\State $R \gets R_{\text{scaled}} S_c^{-1}$

\State \textbf{Output:} $L, R$

\end{algorithmic}
\caption{Construction of low-rank matrices after learning the row/column scaling}
\label{alg:reconstruction_algo}
\end{algorithm*}

\section{Investigating Scaling Matrix $S$ Training Paradigms}
\begin{figure}[!htb]
    \centering
    \includegraphics[width=0.95\linewidth]{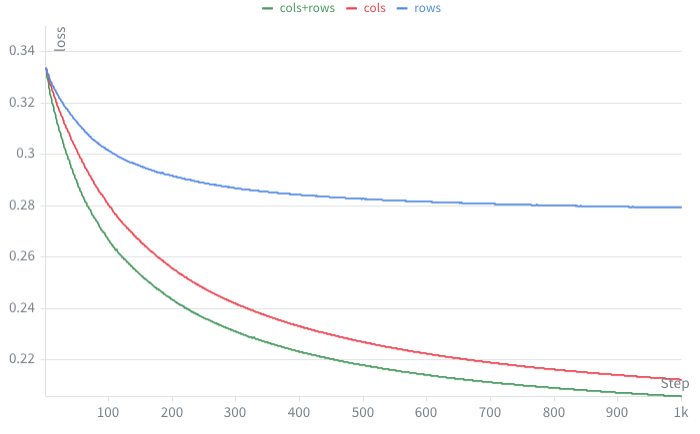}
    \caption{Overview of compression loss when training scaling matrices applied separately and together for rows and columns for Llama 3.1 layer 30 Key matrix}
    \label{fig:appx_row_col_loss_curve_key}
\end{figure}

\begin{figure}[!htb]
    \centering
    \includegraphics[width=1\linewidth]{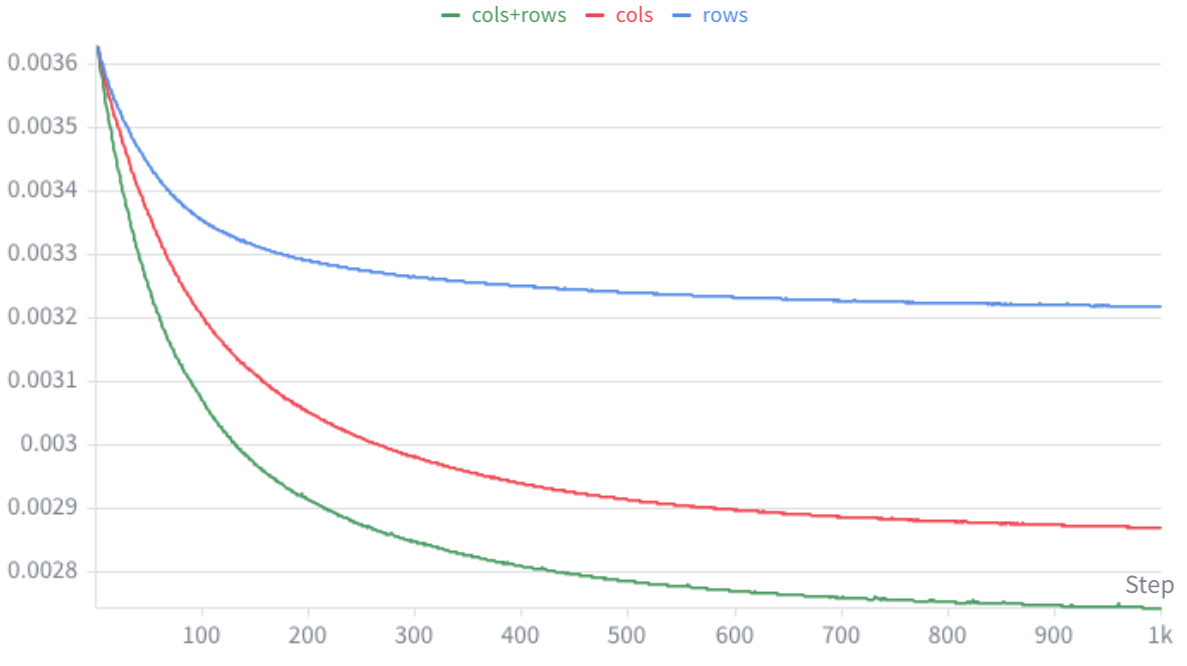}
    \caption{Overview of compression loss when training scaling matrices applied separately and together for rows and columns for Llama 3.1 layer 14 MLP\_down matrix}
    \label{fig:appx_row_col_loss_mlp}
\end{figure}

We conduct additional analysis of the scaling matrix $S$ training to check for isolated and aggregated effects of row and column scaling with respect to the compression loss. For this we use a Llama 3.1 8B model's Key matrix from layer 30 as the test case, see Table \ref{tab:row_col_both_effects} and Figure \ref{fig:appx_row_col_loss_curve_key} and Figure \ref{fig:appx_row_col_loss_mlp} for loss curve of MLP\_down module. The loss curves in Figures \ref{fig:appx_row_col_loss_curve_key} and \ref{fig:appx_row_col_loss_mlp} show a clear benefit of scaling both rows and columns with respect to decreasing the compression loss. Additionally, we execute a test run for training the row and column scaling matrices $S$ separately by training first row and then column scaling, as well as jointly. Table \ref{tab:seq_joint_training_loss} show this result for a single module as an example, we use this information to jointly train all scaling matrices as part of our core methodology.


\begin{table}[!htb]
\centering
\small
\setlength{\tabcolsep}{8pt}
\begin{tabular}{llcc}
\toprule
\textbf{Module} & \textbf{Compression} & \textbf{Loss} & \textbf{Entropy} \\
\midrule

\multirow{4}{*}{L30 Key}
& Baseline       & 0.3340 & 827.88 \\
& Rows           & 0.2780 & 817.00 \\
& Columns        & 0.2120 & 793.00 \\
& Rows + Columns & 0.2060 & 791.79 \\
\midrule

\end{tabular}

\caption{Compression loss and sigma effective rank entropy for different compression strategies after training scaling matrix $S$. Llama 3.1 8B-Instruct at 80\% reduction for layer 30 key matrix}
\label{tab:row_col_both_effects}
\end{table}

\begin{table}[!htb]
\begin{tabular}{lll}
\toprule
               & Loss   \\
\midrule
Sequentially    & 0.215      \\
Jointly & 0.206   \\
\bottomrule
\end{tabular}
\caption{Compression loss for training row and column scaling matrices sequentially (first rows, then columns) and jointly. Llama 3.1 8B-Instr layer 31 Query matrix at 80\% reduction}
\label{tab:seq_joint_training_loss}
\end{table}

\section{Further Analysis on Sigma Values}
\label{appx:sigma_plots}
\begin{figure*}[!htb]
  \includegraphics[
    trim={0cm 0cm 0cm 0cm},
    clip,
    width=0.49\linewidth
  ]{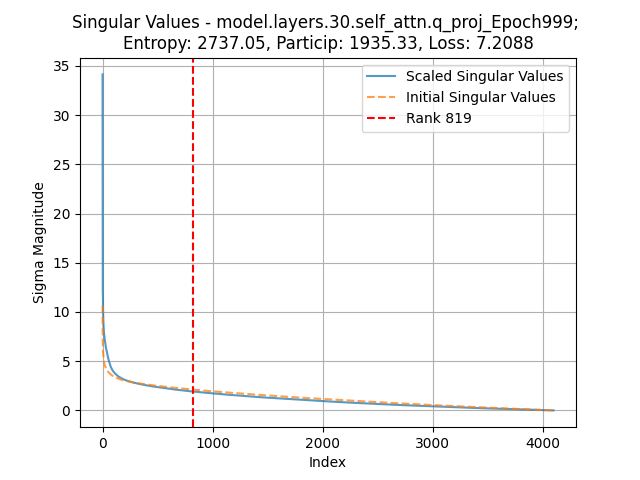}
  \hfill
  \includegraphics[
    trim={0cm 0cm 0cm 0cm},
    clip,
    width=0.49\linewidth
  ]{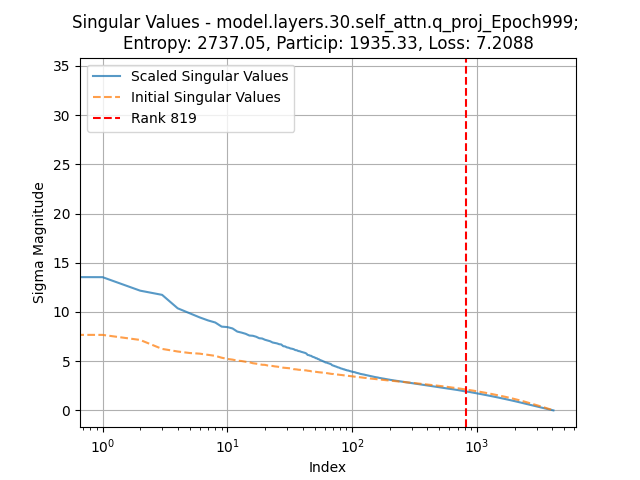}
  \caption{Overview of Llama 3.1 Layer 30 Query matrix. Plots of Sigma values after performing SVD on a scaled and unscaled weight matrix. Side by side comparisons with a logarithmic and linear x axis scaling for an overview of top Sigma values. The dashed Rank line indicates the SVD truncation rank.}
  \label{fig:appx_sigma_plots_l30_query}
\end{figure*}

\begin{figure*}[!htb]
  \includegraphics[width=0.49\linewidth]{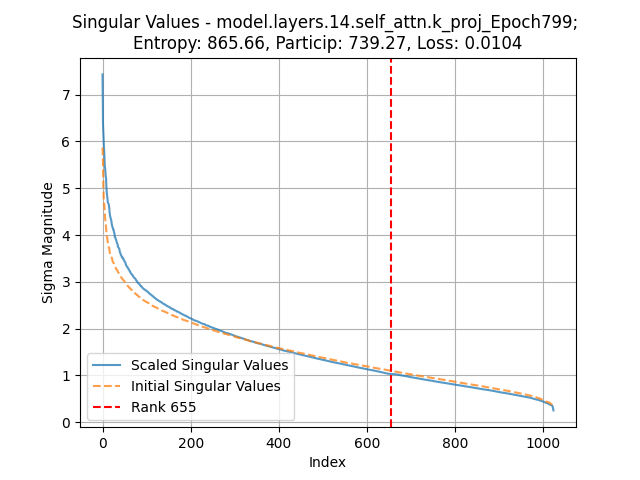} \hfill
  \includegraphics[width=0.49\linewidth]{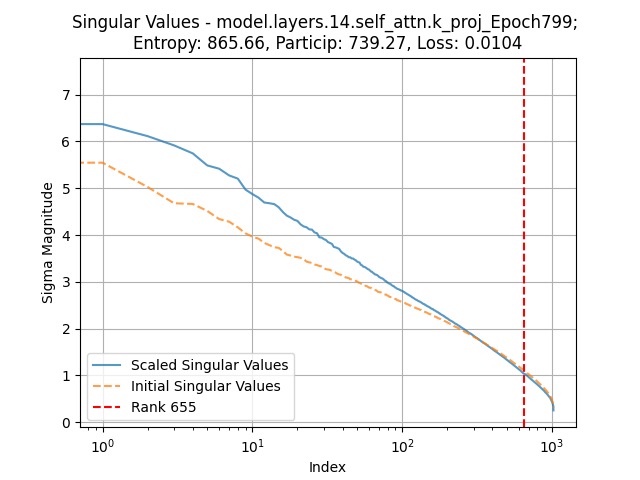}
  \caption {Overview of Llama 3.1 Layer 14 Key matrix. Plots of Sigma values after performing SVD on a scaled and unscaled weight matrix. Side by side comparisons with a logarithmic and linear x axis scaling for an overview of top Sigma values. The dashed Rank line indicates the SVD truncation rank.}
  \label{fig:appx_sigma_plots_l14_key}
\end{figure*}

As described in Section \ref{sect:results}, we expand the analysis of compression loss vs sigma value effective rank entropy in Table \ref{tab:corr_analysis_qwen} for Qwen3-8B model. Additionally see Figures \ref{fig:appx_sigma_plots_l30_query} and \ref{fig:appx_sigma_plots_l14_key} which shows that by applying the scaling matrix $S$ to a weight $W$, there is a downstream effect on the sigma value distribution. The higher end of the sigma values are scaled up, whereas the lower end sees a minor scale-down.

\begin{table}[!htb]
    \small
    \centering
    \begin{tabular}{llll}
    \toprule
        Module & Loss $\Delta$ \% & Entropy $\Delta$ \% & Corr \\ 
        \midrule
        mlp\_down\_proj & -32.935 & -3.468 & 0.877 \\
        mlp\_gate\_proj & -29.183 & -2.874 & 0.882 \\
        mlp\_up\_proj & -28.529 & -2.095 & 0.899 \\ 
        self\_attn\_k\_proj & -53.954 & -7.609 & 0.862 \\
        self\_attn\_o\_proj & -32.383 & -7.797 & 0.868 \\
        self\_attn\_q\_proj & -44.141 & -8.891 & 0.901 \\
        self\_attn\_v\_proj & -36.711 & -4.11 & 0.908 \\
        \bottomrule
    \end{tabular}
    \caption{Overview of loss and effective rank entropy decrease for all seven modules, for Qwen3-8B}
    \label{tab:corr_analysis_qwen}
\end{table}

\section{Codebase and Dataset Links}
\label{sect:codebase_link}
Code repository: \url{github.com/ernlavr/SigmaScale} \\
Custom Alpaca Dataset: \url{https://huggingface.co/datasets/ernlavr/Alpaca-Llama3.1-KD}. We release our contributions under Apache 2.0 license.

\section{Generative AI Disclosure}
As part of this work effort we used generative AI as a coding assistant as well as writing aid for refining text and cross-checking grammar. All generative AI outputs were human cross-checked and validated.

\section{Used Resources Licensing Overview}
\label{appx:licensing}
We summarize licenses of the models and datasets that we have used as part of this study in Table \ref{tab:licenses}

\begin{table*}[h]
\centering
\small
\begin{tabular}{llp{8cm}}
\toprule
\textbf{Dataset / Model} & \textbf{License} & \textbf{Source} \\
\midrule
Alpaca & CC BY-NC 4.0 & \url{https://huggingface.co/datasets/tatsu-lab/alpaca} \\
Llama 3.1 & Custom Meta license & \url{https://github.com/meta-llama/llama-models/blob/main/models/llama3_1/LICENSE} \\
Qwen3 & Apache 2.0 & \url{https://huggingface.co/Qwen/Qwen3-8B} \\
Wikitext & CC BY-SA 3.0 & \url{https://huggingface.co/datasets/Salesforce/wikitext} \\
OpenBookQA & Apache 2.0 & \url{https://github.com/allenai/OpenBookQA/blob/main/LICENSE} \\
ARC-Easy & CC BY-SA 4.0 & \url{https://huggingface.co/datasets/allenai/ai2_arc} \\
WinoGrande & Apache 2.0 & \url{https://github.com/allenai/winogrande/blob/master/LICENSE} \\
PIQA & AFL 3.0 & \url{https://github.com/ybisk/ybisk.github.io/tree/master/piqa} \\
HellaSwag & MIT & \url{https://huggingface.co/datasets/Rowan/hellaswag} \\
\bottomrule
\end{tabular}
\caption{Licensing information for datasets and models used in this study.}
\label{tab:licenses}
\end{table*}

\section{Alpaca Prompt Template}
We define the prompt with a similar template as per original Alpaca dataset. We use Huggingface tokenizer to automatically apply the prompt formatting for Llama and Qwen models. We define overall instructions in the \textit{system} prompt, task-specific instructions and any additional input as the \textit{user} prompt, expected output in the \textit{assistant} section. See Listings \ref{lst:llama_chat_input} and \ref{lst:llama_chat_noinput} for full formatting for Llama 3.1.

\begin{lstlisting}[language={},caption={Llama 3.1 Alpaca-style Chat Template},label={lst:alpaca-llama-template}, label={lst:llama_chat_input}]
<|begin_of_text|><|start_header_id|>system<|end_header_id|>
Below is an instruction that describes a task. Write a response that appropriately completes the request.

<|eot_id|><|start_header_id|>user<|end_header_id|>
### Instruction:
{instruction}

<|eot_id|><|start_header_id|>assistant<|end_header_id|>
{output}
<|eot_id|>
\end{lstlisting}

\begin{lstlisting}[language={},caption={Llama 3.1 Alpaca-style Chat Template},label={lst:alpaca-llama-template}, label={lst:llama_chat_noinput}]
'<|begin_of_text|><|start_header_id|>system<|end_header_id|>
Below is an instruction that describes a task, paired with an input that provides further context. Write a response that appropriately completes the request.

<|eot_id|><|start_header_id|>user<|end_header_id|>
### Instruction: {instruction}

### Input: {input}

<|eot_id|><|start_header_id|>assistant<|end_header_id|>
{output}
<|eot_id|>'
\end{lstlisting}

\end{document}